\documentclass[10pt,twocolumn,journal]{IEEEtran}

\usepackage{epsfig}
\usepackage{cite}
\usepackage{url}
\usepackage{multirow}
\usepackage{epstopdf} 
\usepackage{graphicx}
\usepackage{amsmath, amssymb, amsbsy, algorithmic, algorithm}
\usepackage{booktabs}
\usepackage{balance}
\usepackage{color}
\usepackage{amsbsy}

\pdfoutput=1

\newcommand{\argmax}{\operatornamewithlimits{argmax}}
\newcommand{\argmin}{\operatornamewithlimits{argmin}}

\newlength{\figurewidth}
\newlength{\smallfigurewidth}

\title{Spatial Context based Angular Information Preserving Projection for Hyperspectral Image Classification}

\author{%
Minshan Cui,~\IEEEmembership{Student Member,~IEEE}
and
Saurabh Prasad,~\IEEEmembership{Senior Member,~IEEE}
\thanks{%
The authors are with the Department of Electrical \& 
Computer Engineering, University of Houston, 
Houston, TX, 77004 USA. Contact author: Saurabh Prasad (email: saurabh.prasad@ieee.org).
}
}

\begin{document}

\maketitle
\thispagestyle{empty}
\pagestyle{empty}

\begin{abstract}
Dimensionality reduction is a crucial preprocessing for hyperspectral data analysis - finding an appropriate subspace is often required for subsequent image classification.  In recent work, we proposed supervised angular information based dimensionality reduction methods to find effective subspaces. Since unlabeled data are often more readily available compared to labeled data, we propose an unsupervised projection that finds a lower dimensional subspace where local angular information is preserved. To exploit spatial information from the hyperspectral images, we further extend our unsupervised projection to incorporate spatial contextual information around each pixel in the image. Additionally, we also propose a sparse representation based classifier which is optimized to exploit spatial information during classification - we hence assert that our proposed projection is particularly suitable for classifiers where local similarity and spatial context are both important. Experimental results with two real-world hyperspectral datasets demonstrate that our proposed methods provide a robust classification performance.
\end{abstract}

\section{Introduction}
Hyperspectral imaging techniques have been widely used for a variety of applications such as mineral mapping, surveillance, and environmental monitoring. With its rich spectral information, hyperspectral data provide unique spectral patterns of the materials present in the scene. This capability makes hyperspectral data especially suitable for ground cover classification problems. Although hyperspectral data forms a high dimensional feature space, the information is often effectively represented in a much lower dimensional subspace --- hence, finding an appropriate subspace is crucial for effective hyperspectral data classification. \textcolor{blue}{}.

Several linear and non-linear dimensionality reduction methods have been proposed in literature to address this issue \cite{HN2004,roweis2000nonlinear,tenenbaum2000global,belkin2001laplacian,Sug2007,gao2015discriminative,LSX2010,raducanu2012supervised,zhang2012graph,cui2012novel,hou2010multiple,song2008unified}. The objective function of such dimensionality reduction methods are typically utilize Euclidean distance information. Different from such Euclidean-based dimensionality reduction methods, angle or correlation-based dimensionality reduction methods have been proposed in literature to exploit specific benefits of angular information --- something that is particularly relevant for hyperspectral data. In \cite{ma2007discriminant}, correlation discriminant analysis (CDA) is proposed to transform the data so that the ratio of between-class and within-class sample correlation is maximized. In \cite{cui15angular}, the authors propose several supervised spectral angle based dimensionality reduction methods for robust hyperspectral data classification.     

In typical remote sensing data classification applications, the process of collecting labeled training data is often very time consuming and expensive. Unlabeled data on the other hand is easily available and hence unsupervised dimensionality reduction methods that effectively learn the most appropriate subspace can hence be readily utilized. With that in mind, we propose an unsupervised counterpart of our recently proposed supervised subspace algorithm, the angular discriminant Analysis (ADA).  This unsupervised counterpart is referred to as local similarity preserving projection (LSPP) in this paper. Unlike ADA which requires labeled training samples, LSPP does not require labeled training samples to learn the projection matrix. Additionally, it is well-known that utilizing spatial information in hyperspectral data can dramatically improve the classification accuracies because any such method accounts for the spatial variability of spectral content in local spatial neighborhoods. This follows from the observation that spatially neighboring pixels are highly likely to belong to the same class and have similar spectral characteristics. To incorporate such spatial information of hyperspectral data into our unsupervised projection, we develop a spatial information driven variant of LSPP (SLSPP) which effectively uses the spatial contextual information around each pixel in hyperspectral images to learn the \emph{optimal projection}. 

Recently, sparse representation-based classification (SRC) has been proposed for face recognition \cite{WYG2009} and achieved a great amount of attention in various other applications such as digit recognition \cite{labusch2008simple}, speech recognition \cite{gemmeke2011exemplar}, gesture recognition \cite{zhou2013kernel}, vehicle classification \cite{mei2011robust} as well as remote sensing data classification \cite{MStgars2013,CNT2011,srinivas2011exploiting}. The central idea of SRC is to represent a test sample as a sparse linear combination of atoms in a dictionary which is formed by all of the available training samples. The class label of the test sample is assigned to a class whose reconstruction error is the lowest. In \cite{CNT2011}, the authors propose a join sparsity model to incorporate the contextual information of test samples to improve the classification performance of SRC. However, the contextual information of training samples have not been used. Besides the proposed dimensionality reduction methods, we also propose a sparse representation based classifier which takes into account the spatial information for both training and test samples in this work. Experimental results through two real-world hyperspectral datasets demonstrate that our proposed methods outperform other existing approaches. 

The rest of the paper is organized as follows. In Sec.~\ref{sec:dim_tra}, we briefly introduce the recently developed supervised dimensionality reduction methods (ADA and LADA), and Sec.~\ref{sec:dim_pro1} and Sec.~\ref{sec:dim_pro2} describe the proposed LSPP and SLSPP dimensionality reduction methods respectively. Sec.~\ref{sec:classifier} describes the proposed classification method. In Sec.~\ref{sec:data}, the two practical hyperspectral benchmarking datasets employed to validate the proposed method are described, and the experimental results illustrating the benefits of the proposed work is described in Sec.~\ref{sec:exp}.

\section{Spectral Angle-Based Dimensionality Reduction}
\subsection{ADA and LADA}
\label{sec:dim_tra}
Recently, supervised angular discriminant analysis (ADA) was proposed in \cite{cui15angular}. ADA finds a subspace where the angular separation of between-class samples is maximized while the angular separation of within-class samples is simultaneously minimized. It is mainly used to improve the classification performance of NN with cosine angle distance and SRC with OMP as the recovery method.

Let us define $\{\boldsymbol{x}_{i} \in \mathbb{R}^{d}, y_i \in \{1,2, \ldots, c\}, i=1,2, \ldots, n\}$ to be the $d$-dimensional $i$-th training sample with an associated class label $y_i$, where $c$ is the number of classes and $n$ is the total number of training samples. $n = \sum_{l=1}^{c}{n_{l}}$ where $n_{l}$ denotes the number of training samples from class $l$. Let $\boldsymbol{\mathit{T}} \in \mathbb{R}^{d \times r}$ be the projection matrix, where $r$ denotes the reduced dimensionality. The projection $\mathit{T}_{\textit{ADA}}$ in ADA can obtained by solving
\begin{align}
	\boldsymbol{\mathit{T}}_{\textit{ADA}} 
	&\approx \argmin_{\boldsymbol{\mathit{T}}\in \mathbb{R}^{d \times r}}\left[\operatorname{tr}\big((\boldsymbol{\mathit{T}}^{t}\boldsymbol{\mathit{O}}^{(\text{w})}\boldsymbol{\mathit{T}})^{-1}\boldsymbol{\mathit{T}}^{t}\boldsymbol{\mathit{O}}^{(\text{b})}\boldsymbol{\mathit{T}}\big)\right].
	\label{eq:ada}
\end{align}
where $\boldsymbol{\mathit{O}}^{(\text{w})}$ and $\boldsymbol{\mathit{O}}^{(\text{b})}$ are defined as
\begin{align}
	\boldsymbol{\mathit{O}}^{(\text{w})} &= \sum^{c}_{l=1}\sum^{}_{i:y_i=l} \boldsymbol{\mu}_{l}\boldsymbol{x}_{i}^{t},
	\label{eq:Ow}
	\\
	\boldsymbol{\mathit{O}}^{(\text{b})} &= \sum^{c}_{l=1}n_l\boldsymbol{\mu}\boldsymbol{\mu}_{l}^{t}.
	\label{eq:Ob}
\end{align}  
The problem in \eqref{eq:ada} can be converted to a generalized eigenvalue problem.

The authors in \cite{cui15angular} also extend ADA into its localized variant named local angular discriminant analysis (LADA) by incorporating a locality preserving property through an affinity matrix. LADA was shown to handle more complex, potentially multimodal data distributions. The projection matrix $\mathit{T}_{\textit{LADA}}$ of LADA can be obtained by solving
\begin{align}
	\boldsymbol{\mathit{T}}_{\textit{LADA}} &=  \argmin_{\boldsymbol{\mathit{T}}\in \mathbb{R}^{d \times r}}\left[\operatorname{tr}\big((\boldsymbol{\mathit{T}}^{t}\boldsymbol{\mathit{O}}^{(\text{lw})}\boldsymbol{\mathit{T}})^{-1}\boldsymbol{\mathit{T}}^{t}\boldsymbol{\mathit{O}}^{(\text{lb})}\boldsymbol{\mathit{T}}\big)\right].
	\label{eq:lspdr}
\end{align}

The within-class and between-class outer product matrices of LADA are defined as 
\begin{align}
	\boldsymbol{\mathit{O}}^{(\text{lw})} &= \sum_{i,j=1}^{n}\mathit{W}_{ij}^{(\text{lw})}\boldsymbol{x}_{i}\boldsymbol{x}_{j}^{t}, \label{eq:lada_ow} \\
	\boldsymbol{\mathit{O}}^{(\text{lb})} &= \sum_{i,j=1}^{n}\mathit{W}_{ij}^{(\text{lb})}\boldsymbol{x}_{i}\boldsymbol{x}_{j}^{t},
	\label{eq:lada_ob}
\end{align}
where the weight matrices are defined based on heat kernel $\mathit{A}_{ij}$ as 
\begin{align}
	\mathit{W}^{(\text{lw})}_{ij}&=
	\begin{cases}
		\mathit{A}_{ij}/n_l, & \text{if $y_i,y_j=l$}, \\
		0, & \text{if $y_i\ne y_j$},
	\end{cases}
	\label{eq:lada_w} \\
	\mathit{W}^{(\text{lb})}_{ij}&=
	\begin{cases}
		\mathit{A}_{ij}(1/n-1/n_l), & \text{if $y_i,y_j=l$}, \\
		1/n, & \text{if $y_i\ne y_j$}.
	\end{cases}
	\label{eq:lada_b}
\end{align}

\subsection{LSPP}
\label{sec:dim_pro1}
In this paper, we seek to make two related contributions within the context of angular discriminant analysis --- (1) developing an unsupervised approach to spectral angle based subspace learning, where local spectral angles are preserved following this unsupervised project, and (2) developing a projection that incorporates spatial information when learning such an \emph{optimal} projection. We first form a unsupervised version of ADA which we refer to as local similarity preserving projection (LSPP). It seeks a lower dimensional space where the correlation or angular relationship {between samples that are neighbors in the feature space} are preserved. We can also think of it as an angular equivalent of the commonly employed locality preserving projection (LPP) \cite{HN2004}. 

Let $\textit{x}_i$ be the $i$-th training sample and $\textit{P}$ be the $d \times r$ projection matrix, where $r$ is the reduced dimensionality. The objective function of LSPP can be simplified as
\begin{align}
	\textit{I} &= \sum_{ij}\textit{W}_{ij}(\textit{P}^t\textit{x}_i)^t(\textit{P}^t\textit{x}_j) \nonumber \\
	&= \sum_{ij}\operatorname{\emph{tr}}\big[\textit{W}_{ij}(\textit{P}^t\textit{x}_i)^t(\textit{P}^t\textit{x}_j)\big] \nonumber \\
	&=
	\sum_{ij}\operatorname{\emph{tr}}\big[\textit{W}_{ij}\textit{P}^t\textit{x}_j(\textit{P}^t\textit{x}_i)^t \big] \nonumber \\
	&=
	\sum_{ij}\operatorname{\emph{tr}}\big[\textit{P}^t\textit{W}_{ij}\textit{x}_i\textit{x}_j^t\textit{P}\big] \nonumber \\
	&=
	\operatorname{\emph{tr}}\big[\textit{P}^t\textit{X}\textit{W}\textit{X}^t\textit{P}\big]
\end{align}
The heat kernel $\textit{W}_{ij}\in [0,1]$ between $\textit{x}_i$ and $\textit{x}_j$ is defined as
\begin{equation}
	\textit{W}_{ij}=\exp\left(-\frac{\Vert \textit{x}_i-\textit{x}_j\Vert^2}{\sigma}\right) ,
	\label{eq:aff1}
\end{equation}  
where $\sigma$ is the parameter in the heat kernel.

We impose a constraint ($\textit{P}^t\textit{X}\textit{D}\textit{X}^t\textit{P}=1$ where $D_{ii}=\sum_j\textit{W}_{ij}$) {to avoid biases caused by different samples}. {The larger the value $D_{ii}$ which is corresponding to $i$-th training sample, the more important $i$-th training sample is.}

The final objective function is defined as
\begin{equation}
	\argmax_{\textit{P}}{\ \operatorname{\emph{tr}}\big[\textit{P}^t\textit{X}\textit{W}\textit{X}^t\textit{P}}\big] \qquad \operatorname{s. t.} \quad \textit{P}^t\textit{X}\textit{D}\textit{X}^t\textit{P}=I
	\label{eq:obj1} 
\end{equation}

The problem in \ref{eq:obj1} can be solved as a generalized eigenvalue problem as
\begin{equation}
	\textit{X}\textit{W}\textit{X}^t\textit{P} = \lambda\textit{X}\textit{D}\textit{X}^t\textit{P}
	\label{eq:eig}
\end{equation}
The projection matrix $\textit{P}$ are the eigenvectors corresponding to the $r$ largest eigenvalues.

\subsection{SLSPP}
\label{sec:dim_pro2}
It is well understood from many recent works \cite{TBC2009,CNT2011,JBP2012,cui2013locality}, that by aking into account the spatial neighborhood information, hyperspectral image classification accuracy can be significantly increased. This is based on the observation that spatially neighboring samples in hyperspectral data often consist of similar materials,and hence they are spectrally correlated. In order to utilize spatially neighboring samples in a lower-dimensional subspace, the spatial neighborhood relationship of hyperspectral data should be preserved. To address this problem, we propose a spatial variant of LSPP (SLSPP) in this work to further improve the classification accuracies. Let $\{\textit{z}_{k}, k \in \Omega_i\}$ be the spatial neighborhood samples around a training sample $\textit{x}_{i}$, then the objective function of SLSPP can be reduced to
\begin{align}
	\textit{I} &= \sum_{i}\sum_{k \in \Omega_i} \textit{W}_{ik}(\textit{P}^t\textit{x}_i)^t(\textit{P}^t\textit{z}_{k}) \nonumber \\
	&= \sum_{i}\sum_{k \in \Omega_i}\operatorname{\emph{tr}}\big[\textit{W}_{ik}(\textit{P}^t\textit{x}_i)^t(\textit{P}^t\textit{z}_{k})\big] \nonumber \\
	&=
	\sum_{i}\sum_{k \in \Omega_i}\operatorname{\emph{tr}}\big[\textit{W}_{ik}\textit{P}^t\textit{z}_{k}(\textit{P}^t\textit{x}_i)^t\big] \nonumber \\
	&=
	\sum_{i}\sum_{k \in \Omega_i}\operatorname{\emph{tr}}\big[\textit{P}^t\textit{W}_{ik}\textit{z}_{k}\textit{x}_i^t\textit{P}\big] \nonumber \\
	&=
	\operatorname{\emph{tr}}\big[\textit{P}^tM\textit{P}\big]
\end{align}
where $M = \sum_{i}\sum_{k \in \Omega_i}\textit{W}_{ik}\textit{z}_{k}\textit{x}_i^t$.

The final objective function is defined as
\begin{equation}
	\argmax_{\textit{P}}{\ \operatorname{\emph{tr}}\big[\textit{P}^tM\textit{P}}\big] \qquad \operatorname{s. t.} \quad \textit{P}^t\textit{P}=I
	\label{eq:obj2} 
\end{equation}
Similar to LSPP, the projection matrix $\textit{P}$ are the eigenvectors corresponding to the $r$ largest eigenvalues. 

Note that both LSPP and SLSPP are unsupervised projections in the sense that they do not require labels when learning the projections. We also note that a projection such as SLSPP is particularly beneficial when the backend classifier utilizes the spatial structure in the spatially neighboring pixels for each test pixel during classification. Towards that end, we next propose a modified sparse representation based classifier that utilizes sparse representations of the entire spatial neighborhood of a test pixel simultaneously when making a decision.

\section{Sparse Representation-Based Classification via SBOMP}
\label{sec:classifier}
\subsection{SBOMP}
The simultaneous orthogonal matching pursuit (SOMP) \cite{tropp2006algorithms} and block orthogonal matching pursuit (BOMP) \cite{eldar2010block} are all variants of orthogonal matching pursuit (OMP) that explore the \emph{block structure} of test samples and training samples respectively. In this work, we effectively combine these two recovery methods and propose SBOMP to explore the block structure of both training and test samples simultaneously. SBOMP is illustrated in Algorithm~\ref{sbomp}. Let $x_t$ be a test sample. Assume $\textit{A}_{i}$ contains spatial neighborhood samples around $x_i$ (inclusive of $x_i$), $S$ contains the spatial neighborhood samples of $x_t$ (inclusive of $x_t$) and $K$ is the sparsity level. {SBOMP estimates the coefficient $\hat{\textit{C}}$ based on the $K$ mostly correlated spatial training samples in $\textit{A}$}.

\begin{algorithm}[htbp]
	\caption{SBOMP}
	\label{sbomp}
	\begin{algorithmic}[1]
		\STATE \textbf{Input:} A spectral-spatial training data $\textit{A} = \{\textit{A}_{i}\}_{i=1}^{n}$, test data $\textit{S}$ and row sparsity level $K$.
		\\\hrulefill
		\STATE Initialize $\textit{R}^{0} = \textit{S}$, $\Lambda^{0} = \emptyset$, and the iteration counter $m = 1$.
		\WHILE{$m \le K$}
		\STATE Update the support set $\Lambda^{m} = \Lambda^{m-1} \cup \lambda$ by solving 
		\begin{equation} 
			\lambda = \argmax_{i=1,2,\ldots, n}\|\textit{A}_i^{t}\textit{R}^{m-1}\|_{2,1}. \nonumber 
		\end{equation}
		\STATE Derive the coefficient matrix $C^{m}$ based on
		\begin{equation*}
			\textit{C}^{m} = \left(\textit{A}^{t}_{\Lambda^{m}}\textit{A}_{\Lambda^{m}}\right)^{-1}\textit{A}^{t}_{\Lambda^{m}}S
		\end{equation*} 
		
		\STATE Update the residual matrix $\textit{R}^{m}$
		\begin{equation*}
			\textit{R}^{m} = \textit{S}-\textit{A}_{\Lambda^{m}}\textit{C}^{m}
		\end{equation*}
		\STATE $m \leftarrow m + 1$
		\ENDWHILE
		\\\hrulefill
		\STATE \textbf{Output:} Coefficient matrix $\hat{\textit{C}} = \textit{C}^{m-1}$.
	\end{algorithmic}
\end{algorithm}

\subsection{Classification via SBOMP}
\begin{figure*}[htbp] 
	\centering
	\begin{tabular}{c}
		\includegraphics[width=13cm]{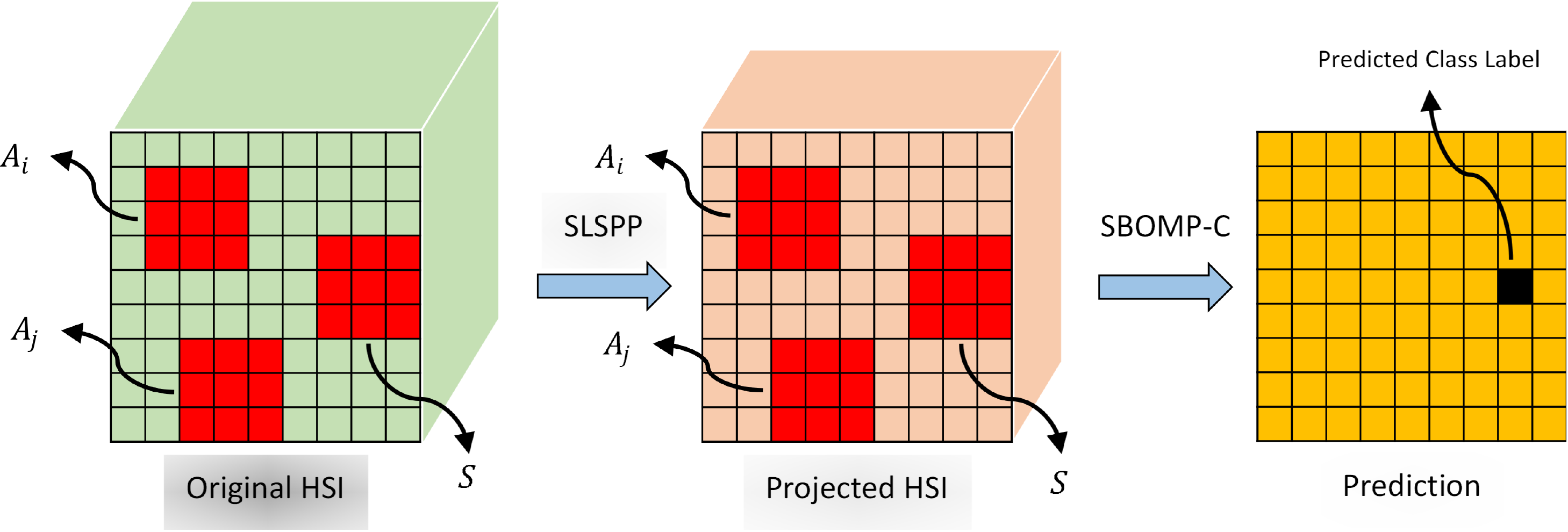} \\
	\end{tabular}
	\caption{Block diagram of the proposed work.}
	\label{fig:diag}
\end{figure*}

The classification method employed after SLSPP is SBOMP-based Classification (SBOMP-C), as described in Algorithm~\ref{sbompc}. Since SLSPP can preserve the spatial neighboring samples for both training and test samples in a lower-dimensional subspace, by using SBOMP-C, we can exploit the block structure relationship effectively in the spatial domain between training and test samples. The block diagram of proposed methods are illustrated in Fig.~\ref{fig:diag}.
\begin{algorithm}[htbp]
	\caption{SBOMP-C}
	\label{sbompc}
	\begin{algorithmic}[1]
		\STATE \textbf{Input:} A spectral-spatial training data $\textit{A} = \{\textit{A}_{i}\}_{i=1}^{n}$, test data $\textit{S}$ and row sparsity level $K$.
		\\\hrulefill
		\STATE Calculate row-sparsity coefficient $\hat{\textit{C}}$ based on \begin{equation*} 
			\hat{\textit{C}} = \text{SBOMP} \ (\textit{A}, S, K)
		\end{equation*}
		\STATE Calculate residuals for each class
		\begin{equation*}
			\textit{r}_{k}(\textit{S}) = \|\textit{S}-\textit{A}\delta_{k}(\hat{\textit{C}})\|_2, \quad k = 1,2, \ldots, c
		\end{equation*}
		\STATE Determine the class label of $\textit{S}$ based on 
		\begin{equation*}
			\omega = \argmin_{k = 1,2,\ldots, c}(\textit{r}_{k}(\textit{S})). \nonumber
		\end{equation*}		 
		\\\hrulefill
		\STATE \textbf{Output:} A class label $\omega$.
	\end{algorithmic}
\end{algorithm}

\section{Experimental Hyperspectral Data}
\label{sec:data}
We validate the proposed approaches with two hyperspectral imagery datasets --- a standard benchmarking dataset from the University of Pavia (an urban ground cover classification problem), and a hyperspectral imagery data collected by our lab to study wetland species composition. 

\subsection{University of Pavia Data}
\begin{figure}[htbp] 
	\centering
	\begin{tabular}{cc}
		\includegraphics[height=5cm]{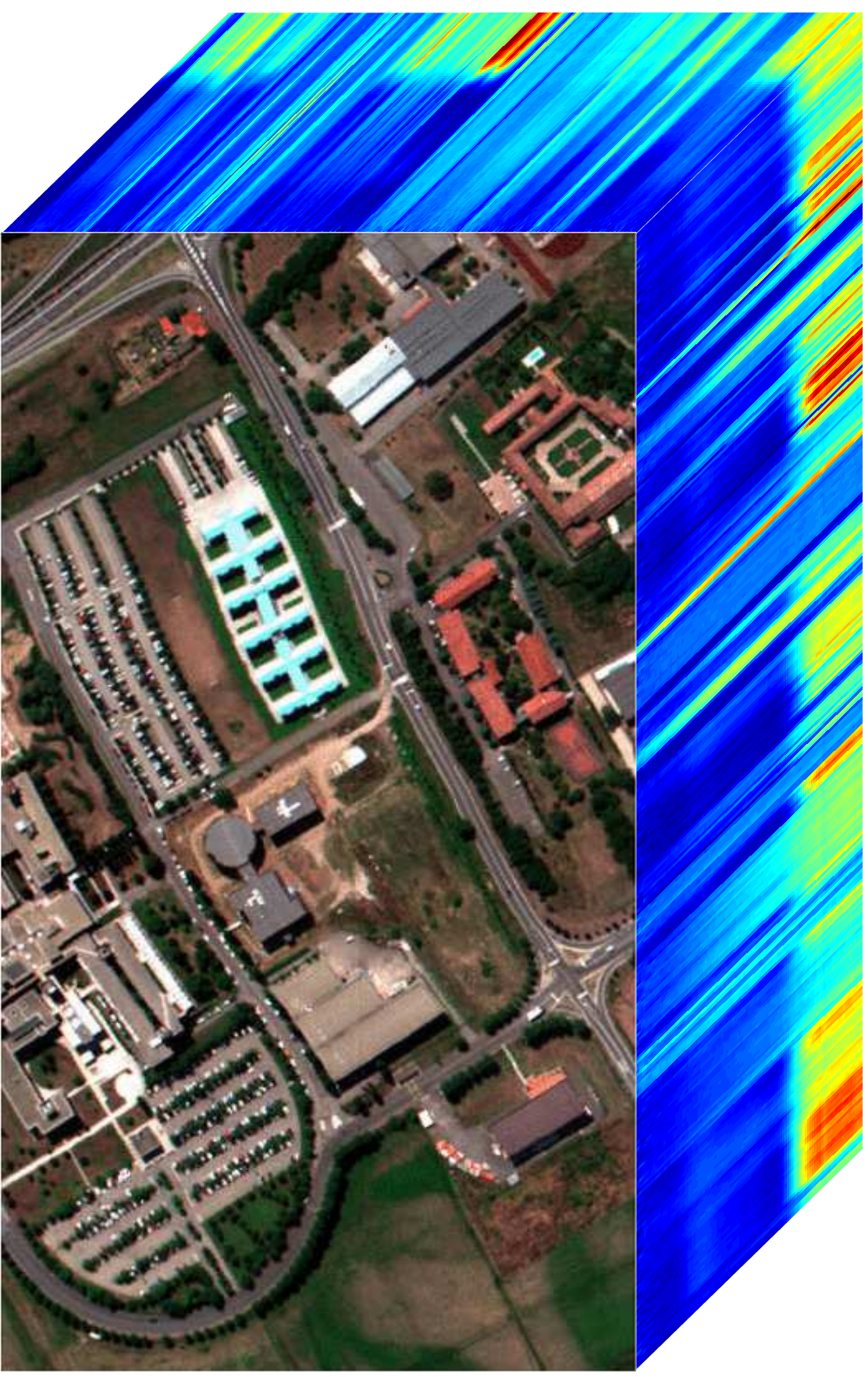} &
		\includegraphics[height=5cm]{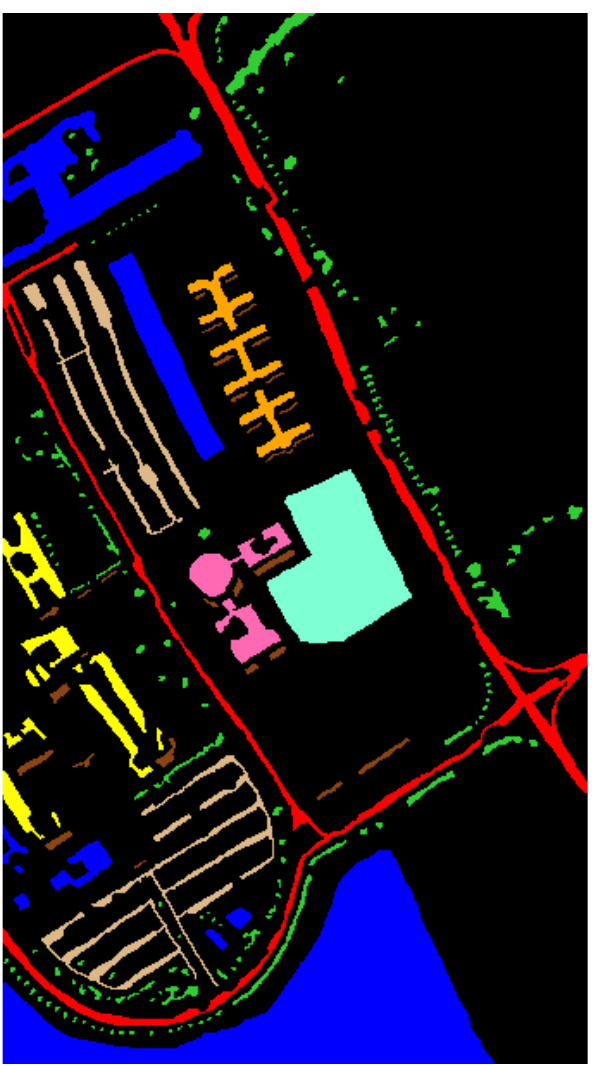} \\
		(a) & (b) \\
	\end{tabular}
	\vspace*{-0.1in}
	\begin{center}
		\includegraphics[height=2cm]{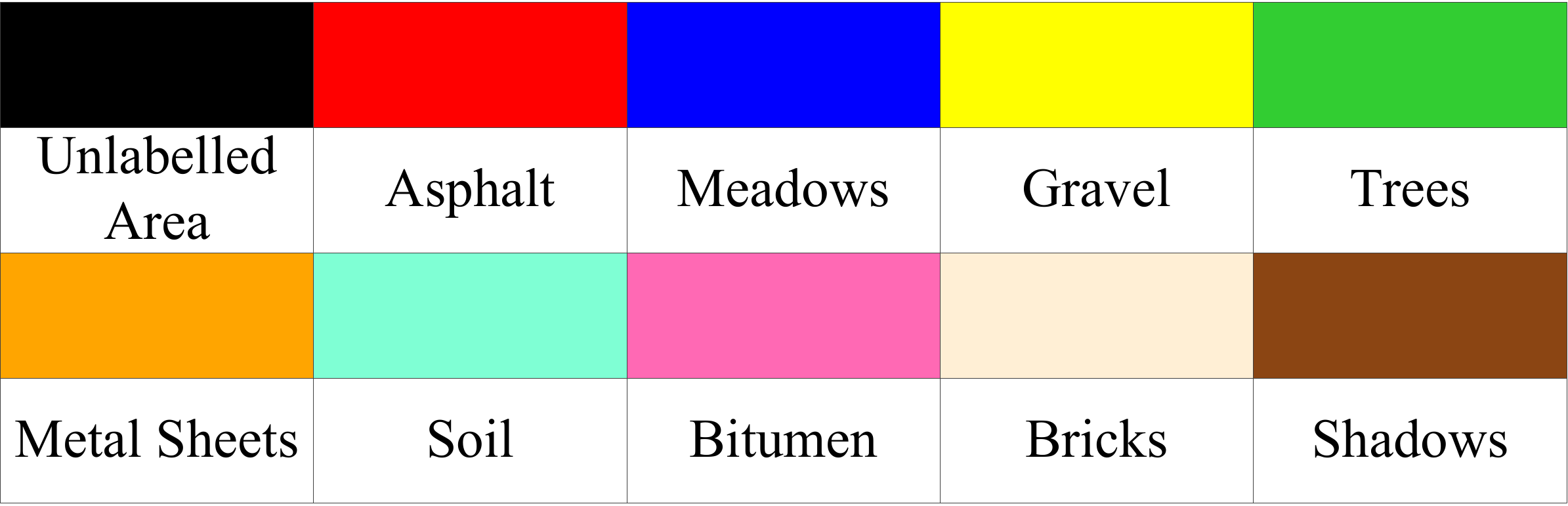} \\
	\end{center}
	\vspace*{-0.1in}
	\caption{
		(a) True color image and (b) ground-truth of the University of Pavia hyperspectral data.}
	\label{fig:data_u}
\end{figure}

The first experimental hyperspectral dataset employed was collected using the Reflective Optics System Imaging Spectrometer (ROSIS) sensor \cite{Gam2004}. This image, covering the University of Pavia, Italy, has 103 spectral bands with a spatial coverage of $610 \times 340$ pixels, and 9 classes of interests are considered in this dataset. A three-band true color image and its ground-truth are shown in Figure \ref{fig:data_u}. This dataset represents a standardized benchmark as it is a well understood and widely studied dataset that is used to quantify performance of machine learning algorithms with hyperspectral data. 

\subsection{Wetland Data}
The second hyperspectral data used in this work was acquired by us in Galveston, Texas in {October, 2014} which includes two different wetland scenes captured at ground-level (side-looking views) over wetlands in Galveston. This is an unprecedented dataset in that it represents ``street view'' type hyperspectral images that have very different characteristics and are very useful for on-ground sensing. The two image cubes are referred to as area 1, and area 2, representing different regions of the wetlands that were imaged --- In addition to common wetland classes, area 2 has Black Mangrove (\emph{Avicennia germinans}) trees in the scene, a species which is of particular interest in ecological studies of wetlands, in addition to Spartina. The images were annotated with species information by an expert in Marine biology, who identified the various species in these hyperspectral scenes. This data was acquired using a Headwall Photonics hyperspectral imager which provides measurements in 325 spectral bands with a spatial size of $1004 \times 5130$. The hyperspectral data uniformly spanned the visible and near-infrared spectrum from $400 nm - 1000 nm$. The objects of interests are primarily vegetation species common in such wetlands. Six different classes were identified in area-1 including soil, \emph{symphyotrichum}, \emph{schoenoplectus}, \emph{spartina patens}, \emph{borrichia} and \emph{rayjacksonia}. The second area includes \emph{Avicennia germinans}, \emph{batis}, \emph{schoenoplectus}, \emph{spartina alterniflora}, \emph{soil}, \emph{water} and \emph{bridge}. Since soil and \emph{schoenoplectus} are included in both areas, the total number of classes in the combined library are eleven. This dataset represents a ``real-world'' application of our method for hyperspectral imaging in the field on arbitrary problems of interest.

\begin{figure*}[htbp] 
	\centering
	\begin{tabular}{c}
		\includegraphics[height=4cm]{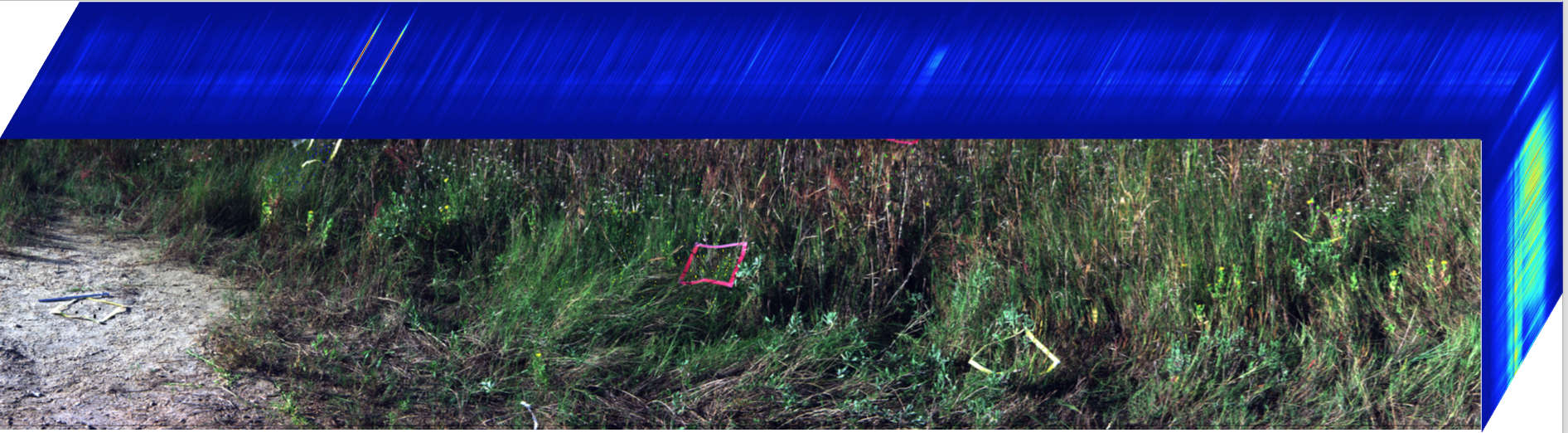} \vspace{-0.2in} \\
		Wetland data, area - 1 \\
		\includegraphics[height=4cm]{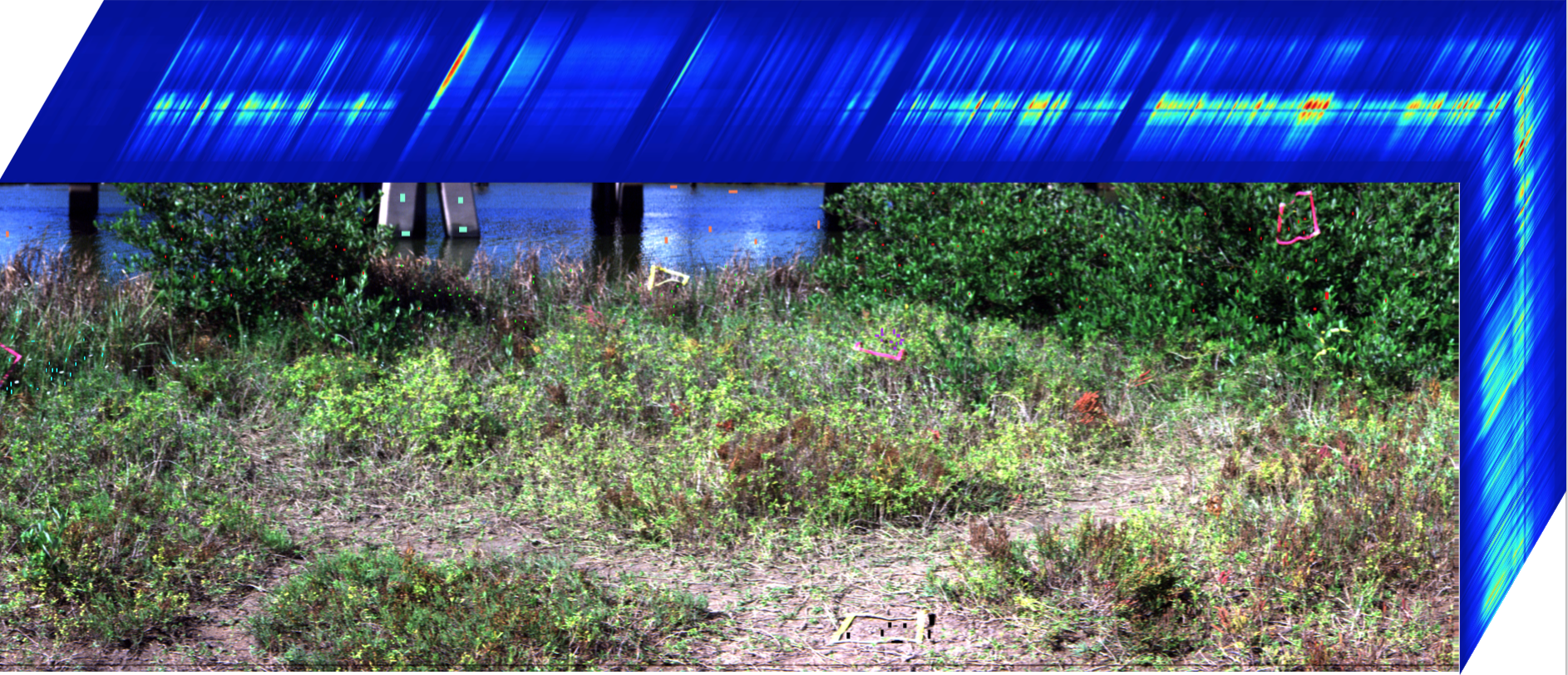} \vspace{-0.2in} \\
		Wetland data, area - 2 \\
	\end{tabular}
	\vspace{-0.1in}
	\begin{center}
		\includegraphics[width=7cm]{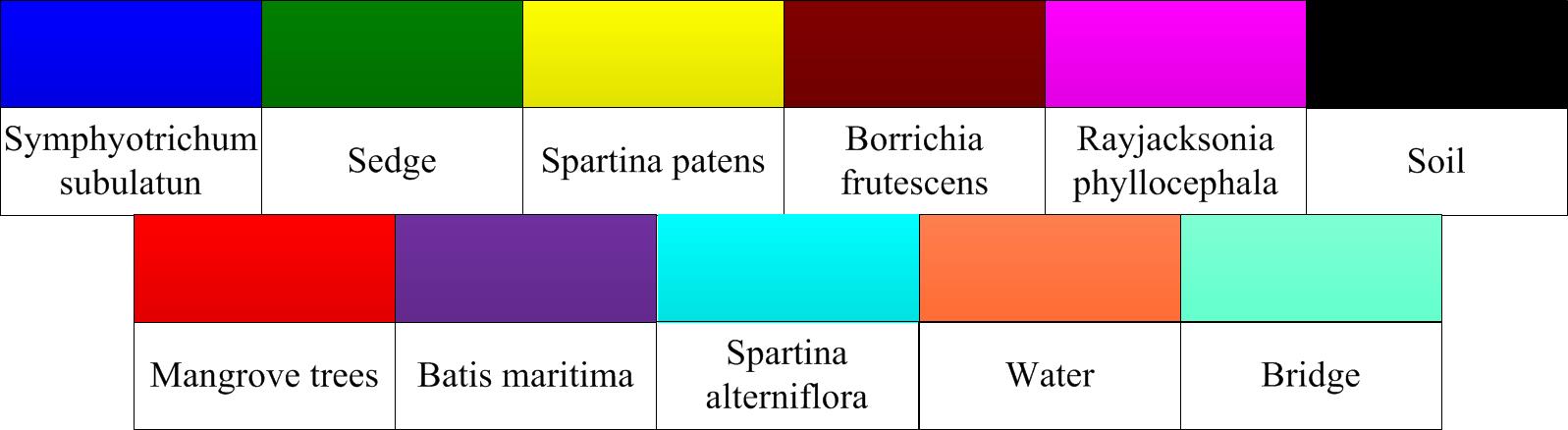}
	\end{center}
	\vspace{-0.2in}
	\caption{True color images of the wetland dataset inset with ground truth.}
\end{figure*}

\section{Experimental Settings and Results}
\label{sec:exp}
The efficacy of the proposed LSPP, SLSPP and SBOMP are evaluated as a function of training samples per class using the two practical hyperspectral datasets mentioned previously. SLSPP--SBOMP-C indicates the data is projected based on SLSPP, and SBOMP-C is employed as the backend classifier. A nearest-neighbor (NN) classifier with cosine angle distance is used after LSPP and LADA projection, since they do not take spatial information into account when deriving the projection matrix. Similar to SLSPP--SOMP-C, LSPP--NN and LADA--NN. Each experiment is repeated 10 times using a repeated random subsampling validation technique, and the average accuracy is reported. The number of test samples per class is fixed to 100 for every random subsampling.

\begin{figure}[h] 
	\centering
	\includegraphics[height=6cm]{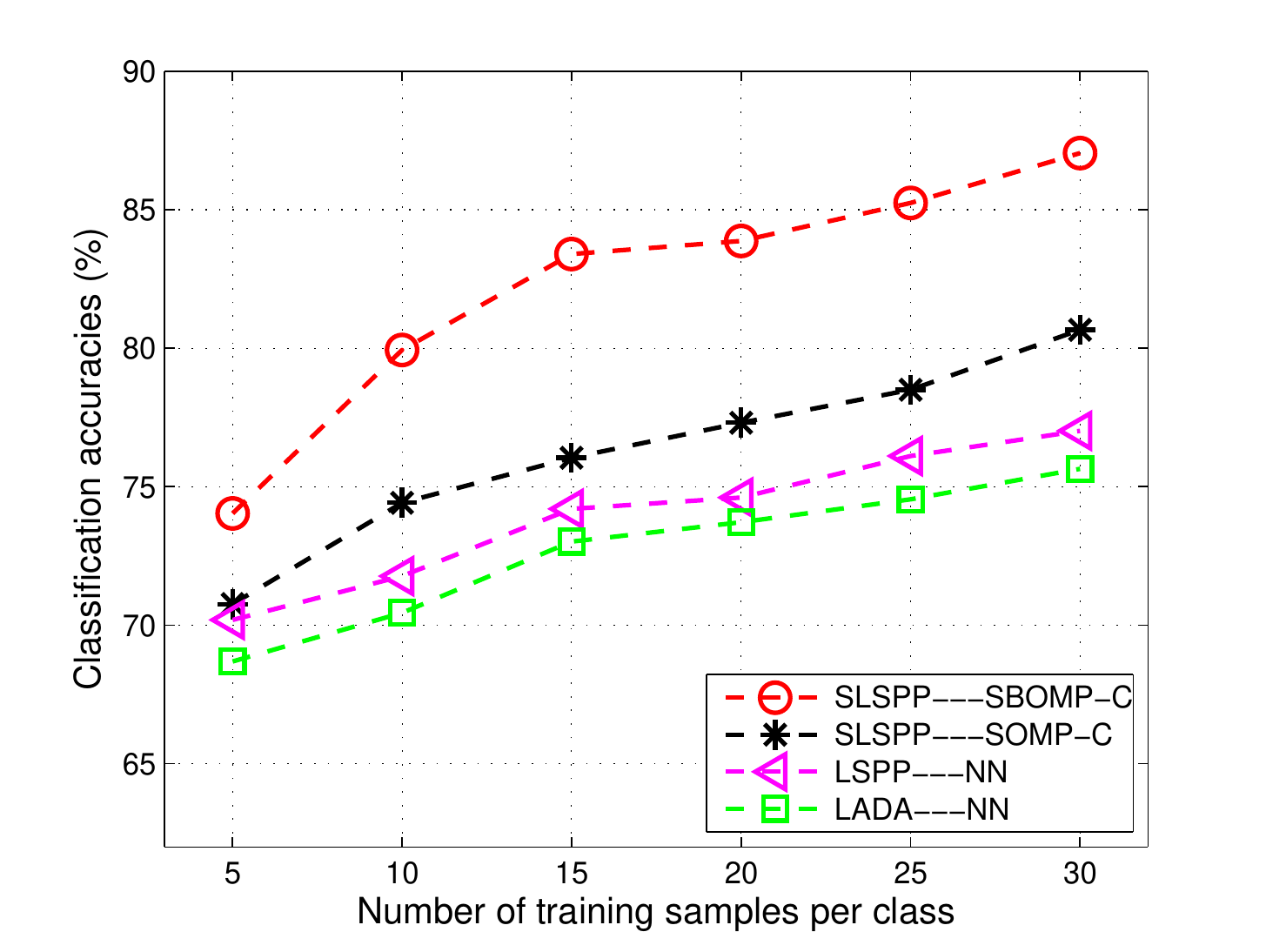} \\
	\caption{Overall classification accuracy (\%) versus number of training samples for the University of Pavia data.}
	\label{fig:result_u}
\end{figure}

Fig.~\ref{fig:result_u} shows the classification accuracies as a function of training samples for the University of Pavia hyperspectral data. The parameter for each algorithm is determined via searching through a wide range of the parameter space and the accuracies reported in this plot is based on the optimal parameter values. As can be seen from this figure, our proposed SLSPP followed by SBOMP-C gives the highest classification accuracies consistently over a wide range of the training sample size. LSPP--NN also gives better classification result compared with LADA--NN. 

\begin{center}
	\begin{figure}[htbp] 
		\centering
		\includegraphics[height=6cm]{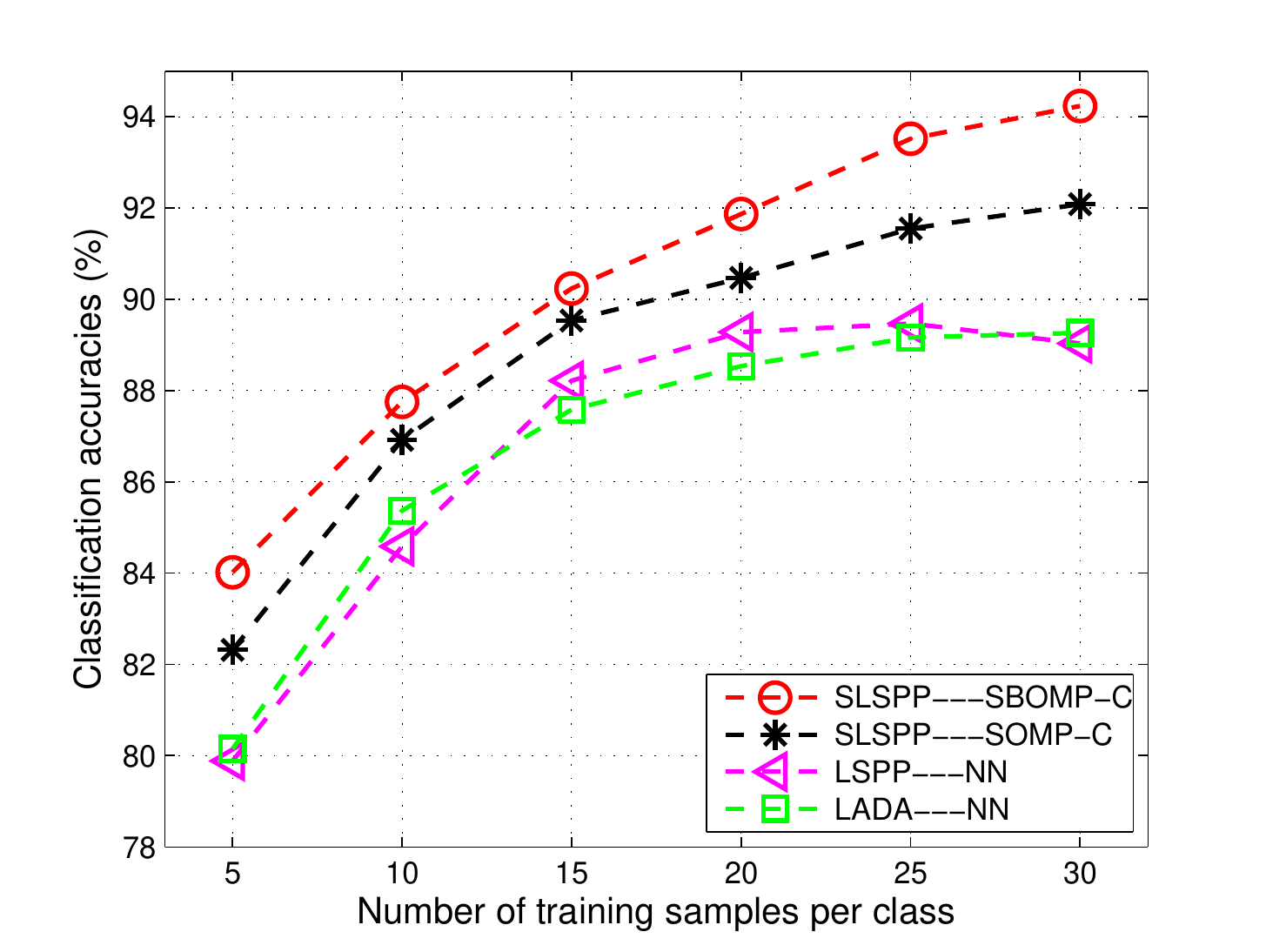} \\
		\caption{Overall classification accuracy (\%) versus number of training samples for the wetland data, area - 1.}
		\label{fig:result_g4}
	\end{figure}
\end{center}

\begin{figure}[htbp]
	\centering
	\includegraphics[height=6cm]{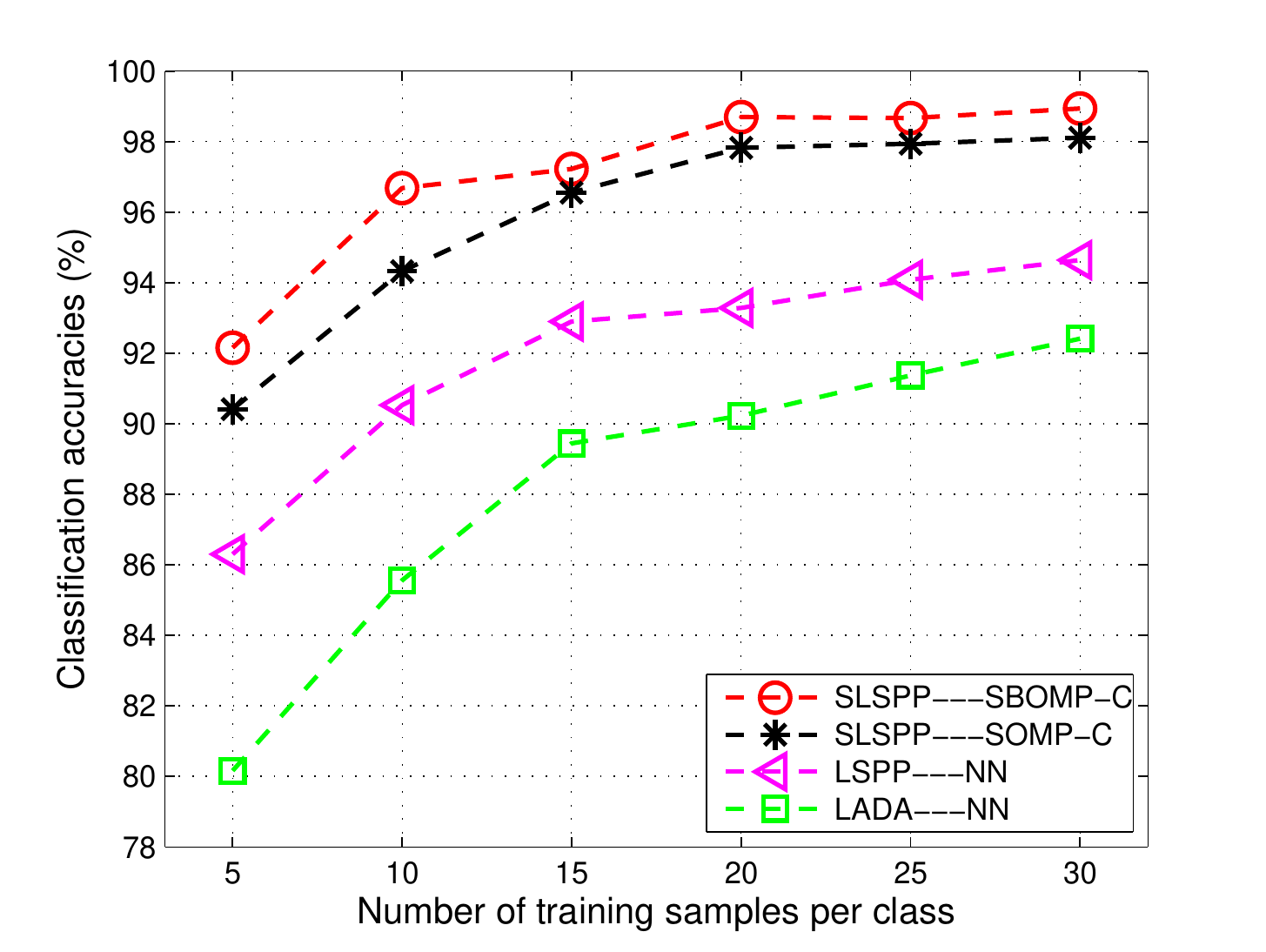} \\
	\caption{Overall classification accuracy (\%) versus number of training samples for the wetland data, area - 2.}
	\label{fig:result_g5}
\end{figure}

We perform a similar analysis using the wetland hyperspectral data. Fig.~\ref{fig:result_g4} and Fig.~\ref{fig:result_g5} plot the classification accuracies with respect to the training sample size per class for wetland area - 1 and wetland area - 2 respectively. It can be seen from the two plots that the proposed methods generally outperform other baseline methods for the vegetation type of hyperspectral data which demonstrate the diverse applicability of the proposed methods.

The class specific accuracies for different datasets are shown in Table~\ref{tab:table1}, Table~\ref{tab:table2} and Table~\ref{tab:table3} respectively. In this experiment, the training sample size per class is fixed to 10 and the test sample size is 100 per class. Each experiment is repeated 10 times and the average accuracy is reported. As can be seen from the class-specific tables for the wetland data, the spatial information plays a crucial role, especially for the species with complex textures and shapes such as \emph{Spartina-Patens}, \emph{Sedge} and \emph{Symphyotrichum}. 
\begin{table*}[htbp]
	\centering
	\caption{Class-specific accuracies (\%) for the University of Pavia Data.}
	\resizebox{13cm}{!}{
		\begin{tabular}{ccccc}
			\toprule
			\textit{\textbf{Class Name / Algorithm}} & \textit{\textbf{LSPP--SBOMP-C}} & \textit{\textbf{LSPP--SOMP-C}} & \textit{\textbf{LSPP--NN}} & \textit{\textbf{LADA--NN}} \\
			\midrule
			\textit{\textbf{Asphalt}} & 77.2  & 60.8  & 31.4  & 30.9 \\
			\textit{\textbf{Meadows}} & 64.4  & 62.5  & 60.2  & 59.6 \\
			\textit{\textbf{Gravel}} & 79.2  & 65.5  & 61.1  & 59 \\
			\textit{\textbf{Trees}} & 88.2  & 79    & 94.4  & 94.1 \\
			\textit{\textbf{Metal Sheets}} & 98.9  & 98.2  & 99.9  & 99.9 \\
			\textit{\textbf{Soil}} & 67.6  & 59.1  & 57.1  & 53.4 \\
			\textit{\textbf{Bitumen}} & 84.9  & 83    & 84.8  & 84.4 \\
			\textit{\textbf{Bricks}} & 60.1  & 64.8  & 66.2  & 62.9 \\
			\textit{\textbf{Shadows}} & 99.2  & 96.9  & 96.9  & 95.9 \\
			\textit{\textbf{Overall Accuracy}} & 80.0  & 74.4  & 72.4  & 71.1 \\
			\bottomrule
		\end{tabular}%
	}
	\label{tab:table1}%
\end{table*}%

\begin{table*}[htbp]
	\centering
	\caption{Class-specific accuracies (\%) for the Wetland, Area-1 data.}
	\resizebox{13cm}{!}{
		\begin{tabular}{ccccc}
			\toprule
			\textit{\textbf{Class Name / Algorithm}} & \textit{\textbf{LSPP--SBOMP-C}} & \textit{\textbf{LSPP--SOMP-C}} & \textit{\textbf{LSPP--NN}} & \textit{\textbf{LADA--NN}} \\
			\midrule
			\textit{\textbf{Soil}} & 99.8  & 100   & 99.8  & 99.9 \\
			\textit{\textbf{Symphyotrichum}} & 86.8  & 85.5  & 78.1  & 76.1 \\
			\textit{\textbf{Sedge}} & 93.4  & 93.3  & 91.6  & 92.1 \\
			\textit{\textbf{Spartina-Paten}} & 68.4  & 59.1  & 64.2  & 56 \\
			\textit{\textbf{Borrichia}} & 91.6  & 91.2  & 91.7  & 93.7 \\
			\textit{\textbf{Rayjacksonia}} & 91.6  & 89    & 84.7  & 92.2 \\
			\textit{\textbf{Overall Accuracy}} & 88.6  & 86.4  & 85.0  & 85.0 \\
			\bottomrule
		\end{tabular}%
	}
	\label{tab:table2}%
\end{table*}%

\begin{table*}[htbp]
	\centering
	\caption{Class-specific accuracies (\%) for the Wetland, Area-2 data.}
	\resizebox{13cm}{!}{
		\begin{tabular}{ccccc}
			\toprule
			\textit{\textbf{Class Name / Algorithm}} & \textit{\textbf{LSPP--SBOMP-C}} & \textit{\textbf{LSPP--SOMP-C}} & \textit{\textbf{LSPP--NN}} & \textit{\textbf{LADA--NN}} \\
			\midrule
			\textit{\textbf{Soil}} & 96.1  & 96.8  & 87.7  & 86.7 \\
			\textit{\textbf{Mangrove Tree}} & 97.1  & 95    & 97.1  & 98.5 \\
			\textit{\textbf{Batis}} & 99.2  & 99.1  & 97.9  & 97.2 \\
			\textit{\textbf{Sedge}} & 97    & 94.3  & 92.3  & 89.8 \\
			\textit{\textbf{Spartina-Alterniflora}} & 89.7  & 90.2  & 75.3  & 48 \\
			\textit{\textbf{Water}} & 99    & 99    & 98.5  & 98.7 \\
			\textit{\textbf{Bridge}} & 99.9  & 99.4  & 94.3  & 85.7 \\
			\textit{\textbf{Overall Accuracy}} & 96.9  & 96.3  & 91.9  & 86.4 \\
			\bottomrule
		\end{tabular}%
	}
	\label{tab:table3}%
\end{table*}%

\begin{figure}[htbp] 
	\centering
	\includegraphics[height=6cm]{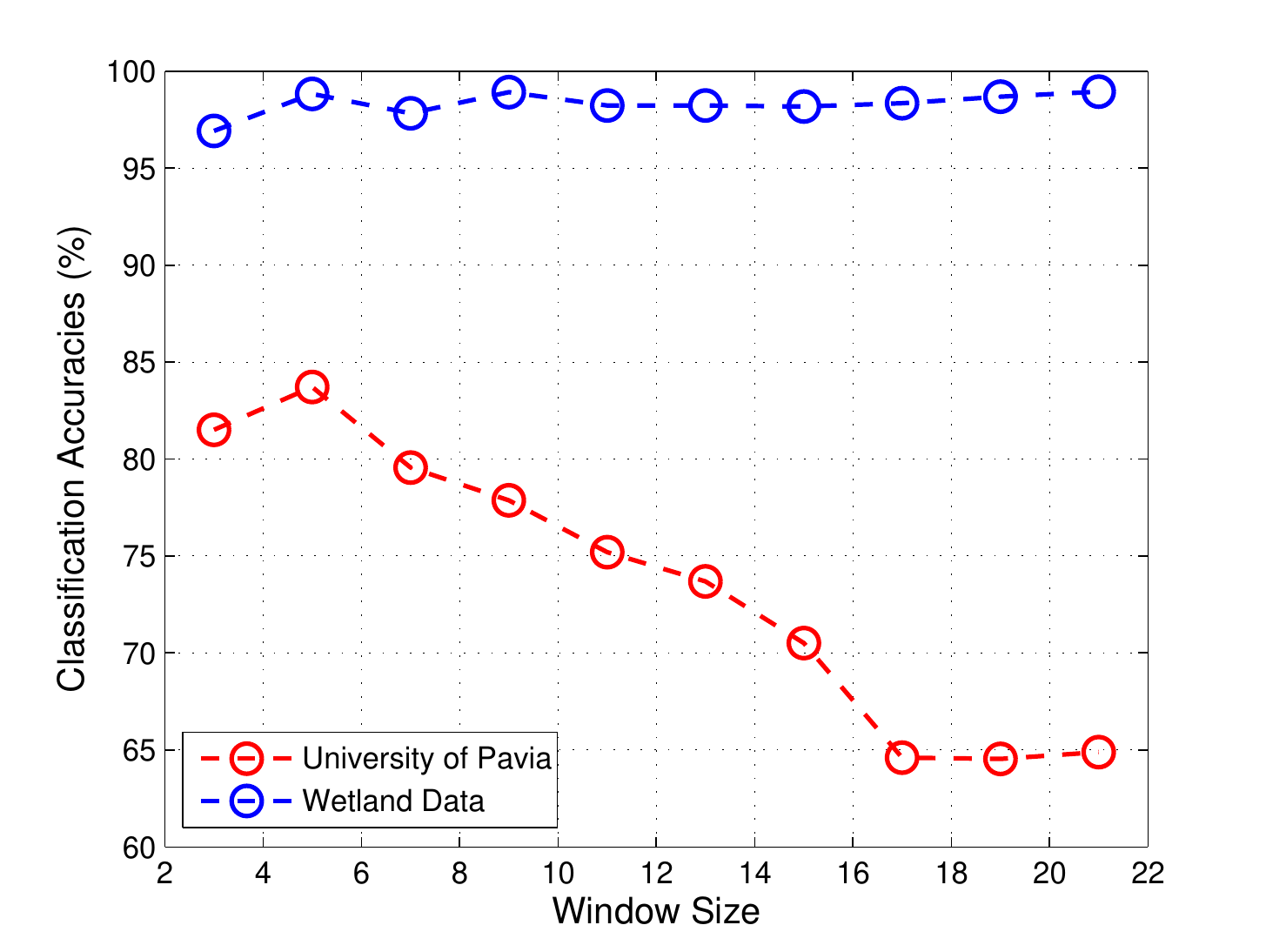} \\
	\caption{Overall classification accuracy (\%) versus different window size for the University of Pavia and Wetland hyperspectral data.}
	\label{fig:win_u}
\end{figure}

Next, we analyze the effect of the window size (that defines the spatial neighborhood) for the SLSPP method. Fig.~\ref{fig:win_u} depicts the classification accuracies as a function of different window size using the University of Pavia dataset. From this figure, we can see that the optimal window size is 5.

\begin{figure}[htbp] 
	\centering
	\begin{tabular}{cc}
		\includegraphics[height=3cm]{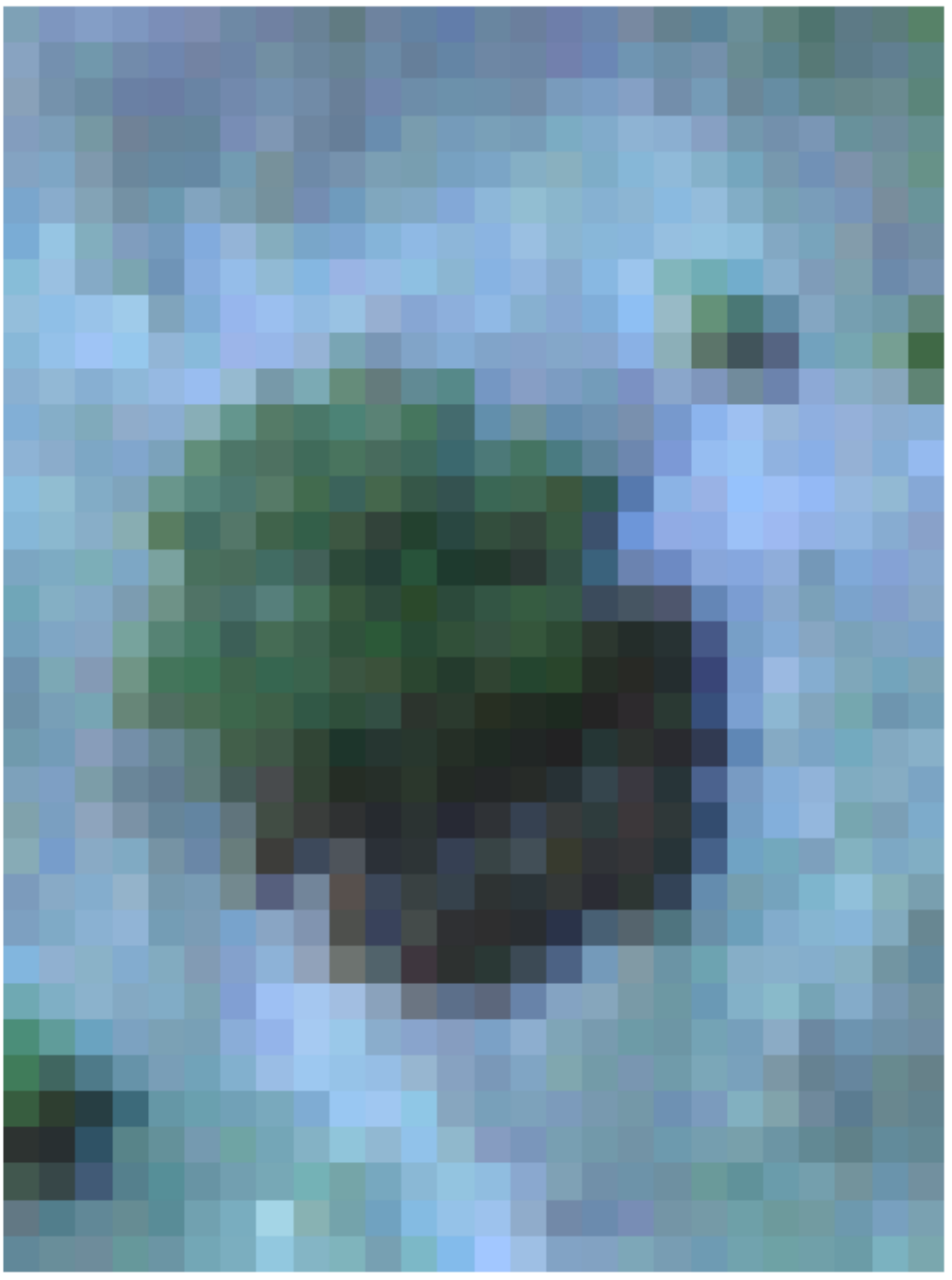} &
		\includegraphics[height=3cm]{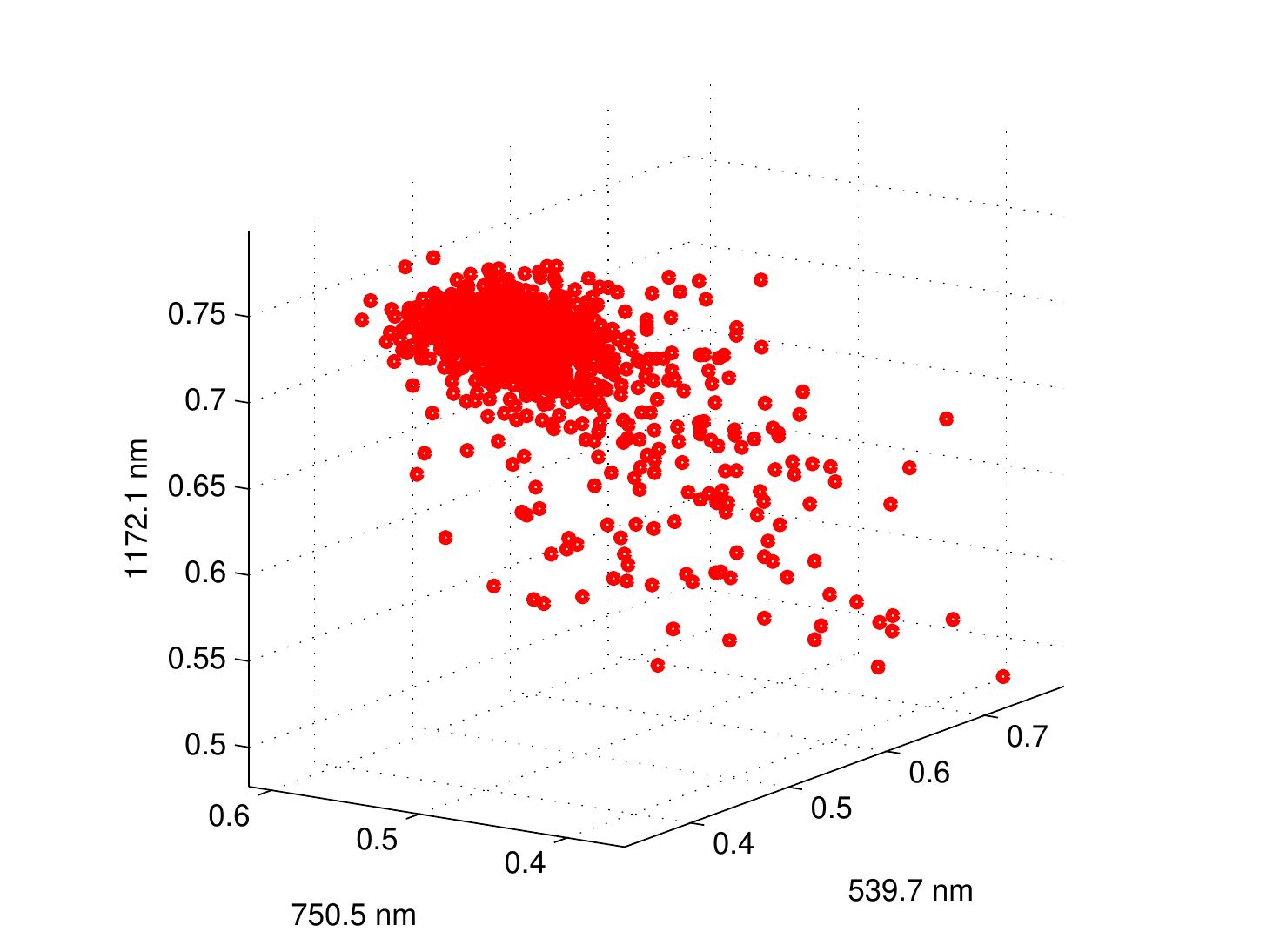} \\
		(a) & (b) \\
		\includegraphics[height=3cm]{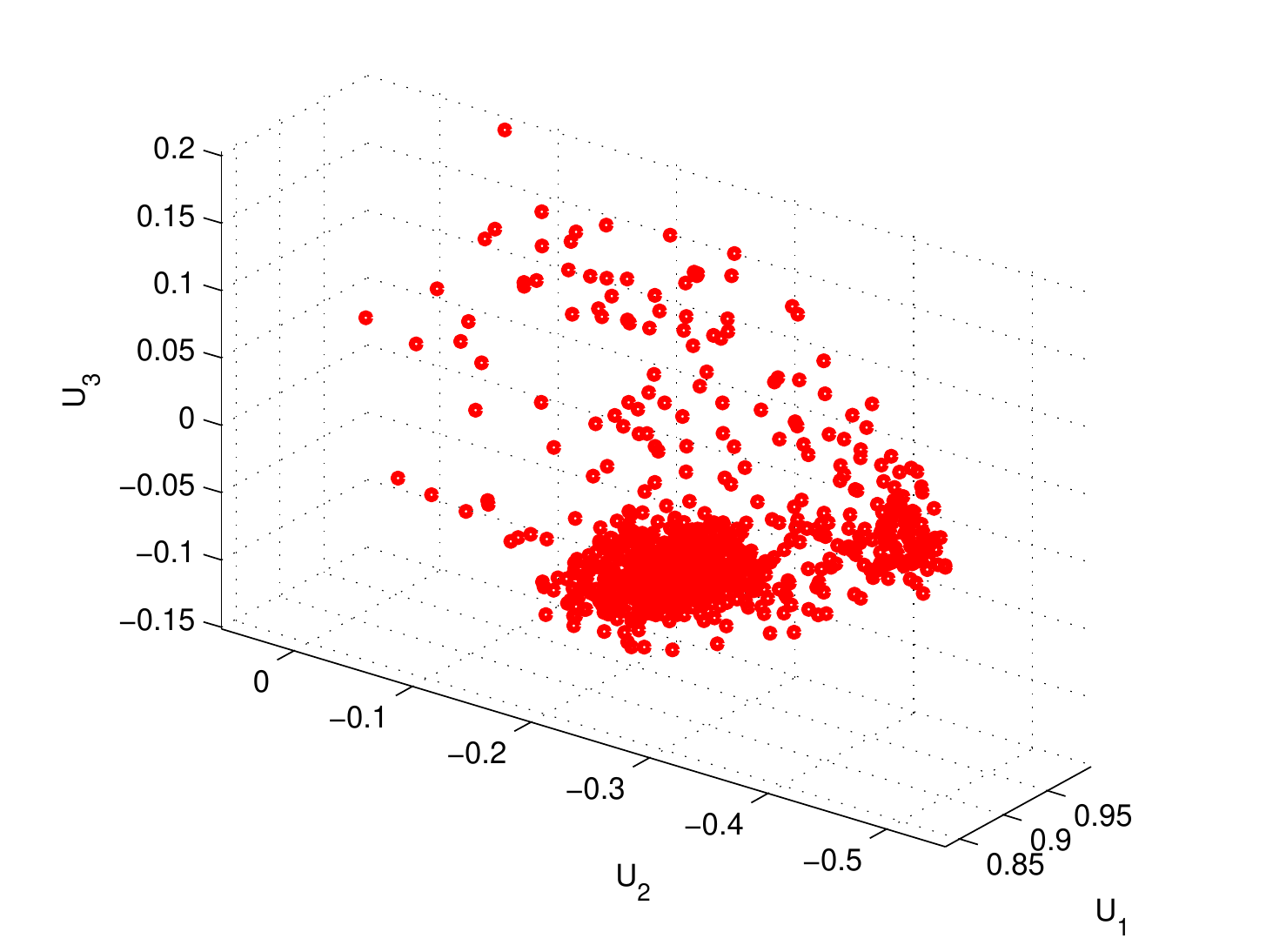} &
		\includegraphics[height=3cm]{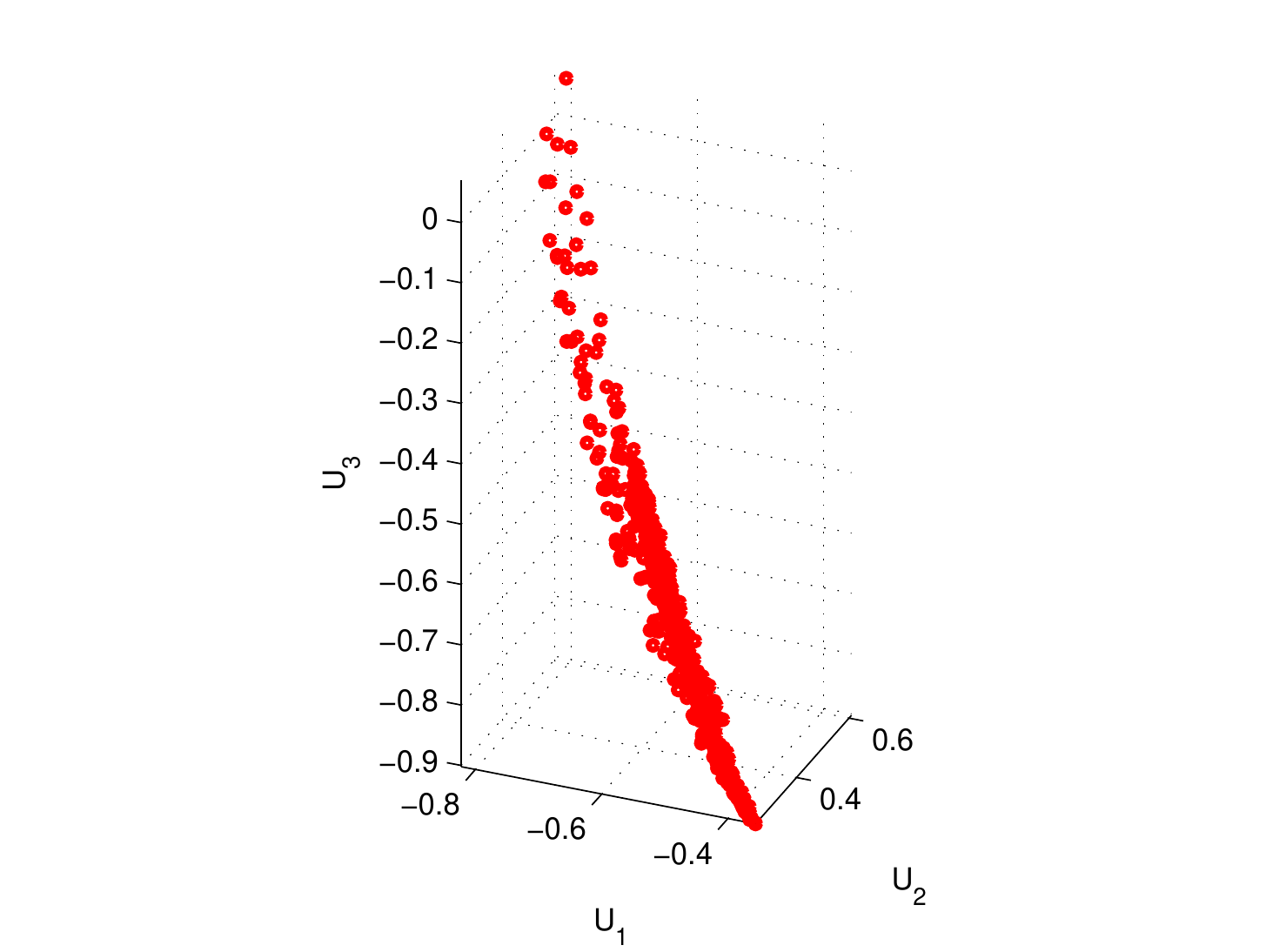} \\
		(c) & (d) \\
	\end{tabular}
	\caption{Illustrating the (a) subset image of the University of Pavia, (b) original samples, (c) SLSPP projected samples and (d) LPP projected samples on the sphere.}
	\label{fig:vis}
\end{figure}

In this work, we also visualize the data distributions after projection for the proposed angle-based SLSPP and the Euclidean-based LPP methods. Figure~\ref{fig:vis} (a) shows the subset image for University of Pavia and Figure~\ref{fig:vis} (b) plots all the training samples used in this experiment on an $\ell_2$ normalized sphere. Figure~\ref{fig:vis} (c) and (d) show the same samples after an SLSPP and LPP projection on an $\ell_2$ normalized sphere respectively. $U_1$, $U_2$ and $U_3$ are the three projections found by SLSPP and LPP corresponding to the largest eigenvalues. As can be seen from this figure, SLSPP is much more effective at preserving the inter-sample relationships in terms of spectral angle in the lower-dimensional subspace compared to LPP.

\section{Conclusion}
In this work, we have presented an unsupervised variant (LSPP) of the recently developed supervised dimensionality reduction method --- ADA, as well as its spatial variant, SLSPP, that utilize spatial information around the samples when learning the projections. By incorporating spatial information in the dimensionality reduction projection, we are able to learn much more effective subspaces that not only preserves angular information among the training pixels in the feature space, but also their spatial neighbors. We also propose a related sparse representation classifier (SBOMP-C) that is optimized for feature spaces generated by SLSPP. Through experimental results based on two practical real-world hyperspectral datasets, we note that the proposed methods significantly outperform the baseline methods.

\bibliographystyle{IEEEtran}
\bibliography{string_def_abbrev,main}

\end{document}